\documentclass[nohyperref]{article}

\usepackage{microtype}
\usepackage{graphicx}
\usepackage{subfigure}
\usepackage{booktabs} 

\usepackage{hyperref}

\usepackage[accepted]{icml2023}

\usepackage{amsmath}
\usepackage{amssymb}
\usepackage{mathtools}
\usepackage{amsthm}
\usepackage{wrapfig}

\usepackage{pifont}

\usepackage[toc,page,header]{appendix}
\usepackage{minitoc}

\usepackage[capitalize,noabbrev]{cleveref}

\theoremstyle{plain}

\theoremstyle{definition}

\theoremstyle{remark}

\usepackage[textsize=tiny]{todonotes}

\icmltitlerunning{Lottery Tickets in Evolutionary Optimization}

\begin{document}

\twocolumn[
\icmltitle{Lottery Tickets in Evolutionary Optimization: \\On Sparse Backpropagation-Free Trainability}




\begin{icmlauthorlist}
\icmlauthor{Robert Tjarko Lange}{tu,scioi}
\icmlauthor{Henning Sprekeler}{tu,scioi}
\end{icmlauthorlist}

\icmlaffiliation{tu}{Technical University Berlin, Berlin, Germany}
\icmlaffiliation{scioi}{Science of Intelligence Cluster of Excellence}

\icmlcorrespondingauthor{Robert Tjarko Lange}{robert.t.lange@tu-berlin.de}

\icmlkeywords{Machine Learning, ICML}

\vskip 0.3in
]



\printAffiliationsAndNotice{} 

\begin{abstract}
Is the lottery ticket phenomenon an idiosyncrasy of gradient-based training or does it generalize to evolutionary optimization? In this paper we establish the existence of highly sparse trainable initializations for evolution strategies (ES) and characterize qualitative differences compared to gradient descent (GD)-based sparse training.
We introduce a novel signal-to-noise iterative pruning procedure, which incorporates loss curvature information into the network pruning step.
This can enable the discovery of even sparser trainable network initializations when using black-box evolution as compared to GD-based optimization.
Furthermore, we find that these initializations encode an inductive bias, which transfers across different ES, related tasks and even to GD-based training.
Finally, we compare the local optima resulting from the different optimization paradigms and sparsity levels.
In contrast to GD, ES explore diverse and flat local optima and do not preserve linear mode connectivity across sparsity levels and independent runs.
The results highlight qualitative differences between evolution and gradient-based learning dynamics, which can be uncovered by the study of iterative pruning procedures.
\end{abstract}

\section{Introduction}

Evolution strategies have recently shown to provide a competitive alternative to gradient-based training of neural networks \citep[e.g.][]{such2017deep, salimans2017evolution}.
Instead of assuming explicit access to gradient evaluations, ES refine the sufficient statistics of a search distribution using information gathered from the black-box evaluation of sampled candidate solutions.
In doing so, modern ES face several scaling challenges: The memory requirements for evaluating populations of large networks quickly become infeasible for common hardware settings. Furthermore, the estimation of the search covariance is often statistically inefficient.
But is it really necessary to evolve full dense networks or can these challenges \textit{in principle} be circumvented by evolving sparse networks? The lottery ticket hypothesis \citep[LTH]{frankle_2019} recently empirically established the existence of sparse network initializations that can be trained to similar performance levels as their dense counterparts.
In this study, we set out to answer whether the existence of such winning initializations is fundamentally tied to gradient-based training or whether sparse trainability can also be achieved in the context of ES.
Furthermore, the LTH has demonstrated yet another application of studying sparse trainability: The empirical analysis of learning dynamics and loss surfaces \citep{frankle_2020b, frankle_2020a}. 
We, therefore, shed light on the differences between gradient descent-based and evolutionary optimization.
\citet{evci2020gradient} previously showed that sparse lottery tickets suffer from reduced gradient flow. GD-based lottery tickets overcome this limitation by biasing the network to retrain to its original dense solution. But does this 'regurgitating ticket interpretation' \citep{maene2021towards} also apply to the setting of gradient-free optimization?
We summarize our contributions as follows:

\begin{figure*}
\centering
\includegraphics[width=0.95\textwidth]{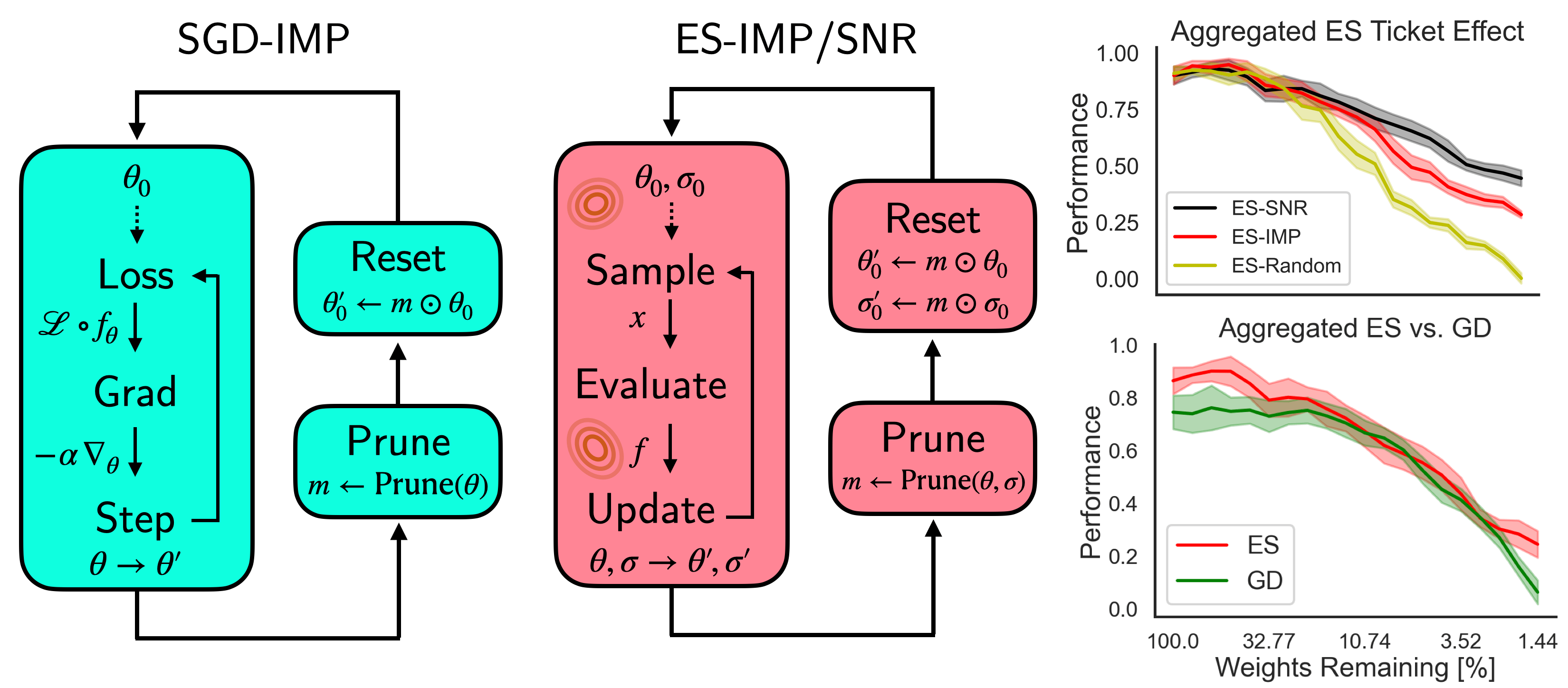}%
\caption{\textbf{Left}. Differences between the iterative magnitude pruning procedures applied to GD and ES training. While GD-based training relies on explicit gradient computation via backpropagation, ES adapts a search distribution based on the principles of biological evolution. In the ES setting, IMP prunes the initial search distribution based on the ratio of the evolved mean magnitude (IMP) and standard deviation (SNR) at each pruning iteration. \textbf{Right, Top}. Task-normalized aggregated ticket effect in ES. IMP-derived lottery ticket initializations outperform random pruning for ES-based optimization. SNR pruning provides an additional sparse trainability improvement. \textbf{Right, Bottom}. The ES ticket procedure yields sparse initializations with performance on par with GD. At high sparsity levels ES tickets outperform GD. The task-specific results are normalized across conditions (random, magnitude, SNR) to lie within $[0, 1]$ range. We average the normalized scores across 12 tasks, 2 different network classes, 3 ES and over 5 independent runs \& plot standard error bars.}
\label{fig:conceptual}
\end{figure*}

\begin{enumerate}
	\item We apply iterative magnitude pruning \citep[][IMP; figure \ref{fig:conceptual} left]{han_2015} to the ES setting and establish the existence of highly sparse evolvable initializations. They consistently exist across different ES, architectures (multi-layer perceptrons/MLP, \& convolutional neural networks/CNN) and tasks (9 control \& 3 vision tasks). The LTH phenomenon does, therefore, not depend on gradient-based training or the implied masked gradient flow (Section \ref{sec3:existence}, Figure \ref{fig:conceptual} right, top).
        \item We introduce a novel iterative pruning procedure based on the signal-to-noise ratio (SNR) of the ES search distribution, which accounts for the loss geometry information encoded by the search covariance. This pruning scheme leads to initializations, which are trainable at higher degrees of sparsity (Figure \ref{fig:conceptual} right, top).
	\item ES tickets outperform sparse random initializations when transferred to related tasks (Section \ref{sec4:transfer}). They can be transferred between different evolutionary strategies and to GD-based training. Hence, ES tickets do not overfit to their training paradigm or task setting. Instead, they capture moderately task-general inductive biases, which can be amortized in different settings.
	\item GD-based lottery tickets are not well suited for training moderately sized neural networks at very high sparsity levels (Section \ref{sec3:existence}, Figure \ref{fig:conceptual}, bottom). For ES-derived lottery tickets, on the other hand, we find that they remain trainable at very high sparsity. For vision-based tasks the performance degradation of GD accelerates and correlates with a strong increase in the sharpness of obtained local optima. ES optima, on the other hand, remain flat for higher sparsity (Section \ref{sec5:analysis}).
	\item While GD-based trained ticket initializations preserve linear mode connectivity at low levels of sparsity, ES tickets do not. Instead, they converge to a diverse set of flat local optima across sparsity levels (Section \ref{sec3:existence}). This questions the 'regurgitation ticket interpretation' and the implicit loss landscape bias of ES-based tickets: ES tickets exist despite the fact that they do not repeatedly converge to the same loss basin. This highlights the potential of ES-based ensemble methods in order to compensate for the expensive pruning procedure. 
\end{enumerate}

\section{Background \& SNR Pruning Procedure}
\label{sec2:background}

\textbf{Iterative Magnitude Pruning in Deep Learning.} Lottery ticket initializations are traditionally found using an expensive iterative pruning procedure \citep[Figure \ref{fig:conceptual}, left; e.g.][]{han_2015, lange2020_lottery_ticket_hypothesis}: Given a dense network initialization $\theta_0$, one trains the parameters using GD.
The final weights $\theta$ are then used to construct a binary mask $m$ based on a simple heuristic: Prune the fraction $p \in (0, 1)$ of smallest magnitude weights. The cutoff threshold is constructed globally and thereby implies a different pruning ratio per layer.
Afterwards, one resets the remaining weights to their original initialization and iterates the procedure using the remaining non-zero weights, $m \odot \theta_0$.
The ticket effect for a given level of sparsity $(1 - p)^k$ can be measured by the performance difference between a sparsity-matched randomly pruned network and the corresponding IMP initialization, $m_k \odot \theta_0$.
Previously, it has been shown that a positive ticket effect can be observed throughout different GD-based training paradigms including computer vision \citep{frankle_2019, frankle_2020a}, reinforcement learning \citep{yu_2019, vischer2021lottery}, natural language processing \citep{chen2020lottery}, graph neural networks \citep{chen2021unified} and generative models \citep{kalibhat2021winning}. 
In contrast to previous work, we investigate whether the lottery ticket phenomenon is an idiosyncratic phenomenon for gradient-based optimization.

\textbf{Evolution Strategies.} ES constitute a set of random search algorithms based on the principles of biological evolution. 
They adapt a parametrized distribution in order to iteratively search for well performing solutions.
After sampling a population of candidates, their fitness is estimated using Monte Carlo (MC) evaluations. The resulting scores are used to update the search distribution.
We focus on two representative ES classes, which have been used in neuroevolution.\\
\underline{\textit{Finite Difference-based ES}}: A subset of ES use random perturbations to MC estimate a finite difference gradient:
\vspace{-0.1cm}
	$$\nabla_\theta \mathbb{E}_{\epsilon \sim \mathcal{N}(0, I)} F(\theta + \sigma \epsilon) = \frac{1}{\sigma} \mathbb{E}_{\epsilon \sim \mathcal{N}(0, I)} [F(\theta + \sigma \epsilon) \epsilon]$$
	This estimate is then used along standard GD-based optimizers to refine the search distribution mean $\theta$ \citep{salimans2017evolution}.
    They differ in their use of fitness shaping, anti-correlated noise, elite selection and the covariance structure. \\
\underline{\textit{Estimation-of-Distribution ES}}: A second class of ES does not rely on noise perturbations or low-dimensional approximations to the fitness gradient. Instead, algorithms such as CMA-ES \citep{hansen2001completely} rely on elite-weighted mean updates and iterative covariance matrix estimation. They aim to shift the search distribution into the direction maximizing the likelihood of fitness improvement.

\textbf{Sparsity \& Pruning in Gradient-Free Optimization.} The NEAT algorithm \citep{stanley2002evolving} co-evolves network architectures and their weights. The connectivity is changed throughout the procedure, which often times naturally leads to the emergence of sparse architectures. \citet{mocanu2018scalable} used an dynamic sparse training algorithm inspired by evolutionary principle to train networks with dynamic topology. Finally, \citet{mocanu_2016} previously investigated the performance of sparse Boltzmann Machines, which do not rely on gradient computation via backpropagation. Here, we investigate the sparse trainability of otherwise static neural networks and the LTH in ES.

\textbf{Searching for Lottery Tickets in Evolution Strategies.} We introduce an ES-generalized iterative pruning procedure and focus on search strategies that adapt the sufficient statistics of a multivariate diagonal Gaussian distribution. 
At each pruning iteration the mean initialization is pruned based on its final estimate (Figure \ref{fig:conceptual}, middle). Furthermore, we consider a signal-to-noise ratio pruning heuristic, which prunes weights based on the ratio of the mean magnitude and the standard deviation of the search distribution, $|\theta|/\boldsymbol{\sigma}$. SNR pruning implicitly incorporates information about the loss geometry/curvature into the sparsification. The procedure is summarized in Algorithm \ref{alg:snr}. We note that \citep{blundell2015weight} previously considered a SNR criterion in the context of zero-shot pruning of Bayesian neural networks.

\begin{algorithm}[tb]
   \caption{SNR-Based Iterative Pruning for ES}
   \label{alg:snr}
\begin{algorithmic}
   \STATE {\bfseries Input:} Pruning ratio $p \in (0, 1)$, iterations $T \in \mathbb{Z}_+$, \texttt{ES} algorithm, $G$ generations, $N$ population size.
   \STATE Initialize the ES search distribution $\theta_0, \boldsymbol{\sigma}_0 \in \mathbb{R}^D$.
   \STATE Initialize a dense pruning mask $m_0 = I_D \in \mathbb{R}^D$.
   \FOR{$t=0$ {\bfseries to} $T - 1$}
   \STATE \# Construct the sparse ES statistics: 
   \STATE $\theta_{0, t} = m_t \odot \theta_0$ and $\boldsymbol{\sigma}_{0, t} = m_t \odot \boldsymbol{\sigma}_0$.
   \STATE \# Evolve non-pruned weights only:
   \FOR{$g=0$ {\bfseries to} $G-1$}
   \STATE \underline{Sample}: $\mathbf{x}_i \sim \mathcal{N}(\theta_{g, t}, \boldsymbol{\sigma}_{g, t} I_D), \forall i=1,\dots, N$.
   \STATE \underline{Evaluate} candidate fitness: $\mathbf{f}_i, \forall i=1,\dots, N$.
   \STATE \underline{Update} ES: $\theta_{g+1, t}, \boldsymbol{\sigma}_{g+1, t} \leftarrow \texttt{ES}(\theta_{g, t}, \boldsymbol{\sigma}_{g, t}, \mathbf{x}, \mathbf{f})$.
   \ENDFOR
   \STATE Compute sparsity level $s_t = (1 - p)^{t+1}$.
   \STATE Compute SNR threshold $\rho(s_t)$ by sorting $ |\theta_{G, t}|/\boldsymbol{\sigma}_{G, t}$.
   \STATE Construct the next mask $m_{t+1} = \mathbf{1}_{|\theta_{G, t}|/\boldsymbol{\sigma}_{G, t} > \rho_t}$.
   \ENDFOR
\end{algorithmic}
\end{algorithm}

Importantly, pruning for ES reduces the memory requirements due to the smaller dimensionality of the stored mean and covariance estimates. 
While most ES initialize the search mean at zero, we instead sample the initialization according to standard network initializations allowing for an initialization effect. This ensures comparability between GD and ES-based iterative pruning.
Throughout the main text, we mainly focus on 4 ES including PGPE \citep{sehnke2010parameter}, SNES \citep{schaul2011high}, Sep-CMA-ES \citep{ros2008simple} and DES \citep{lange2022discovering}. They all make use of a diagonal covariance, which enables the computation of weight-specific SNRs used to compute pruning thresholds. All ES and GD training algorithms were tuned using the same amount of compute.\footnote{We compare against PPO \citep{schulman_2017} for control tasks and simple cross-entropy minimization with Adam for image classification. We refer the interested reader to the supplementary information (SI) for task-specific hyperparameters.
}

\textbf{Measuring Ticket Effect Contributions.} The ticket effect can be decomposed into three ingredients: The pruning-derived mask, the implied initialization of the remaining non-pruned weights, and the layerwise pruning ratio. \citet{vischer2021lottery} proposed to estimate each contribution using a set of permutation experiments. 
By permuting the non-masked weights at each pruning iteration, one may estimate the contribution of the extracted weight initialization. 
If the performance of the sparsified networks remains unharmed, the weight effect is small. If additionally permuting the binary mask greatly damages the trainability, the mask effect is large.
Finally, comparing a randomly pruned network with the doubly permuted baseline allows us to estimate the impact of the layerwise pruning ratio implied by the iterative pruning. 
In this work, we additionally consider the gap between IMP and SNR pruned network initializations. The difference can be attributed to the information resulting from curvature estimation of the ES covariance.

\begin{figure*}
\centering
\includegraphics[width=0.95\textwidth]{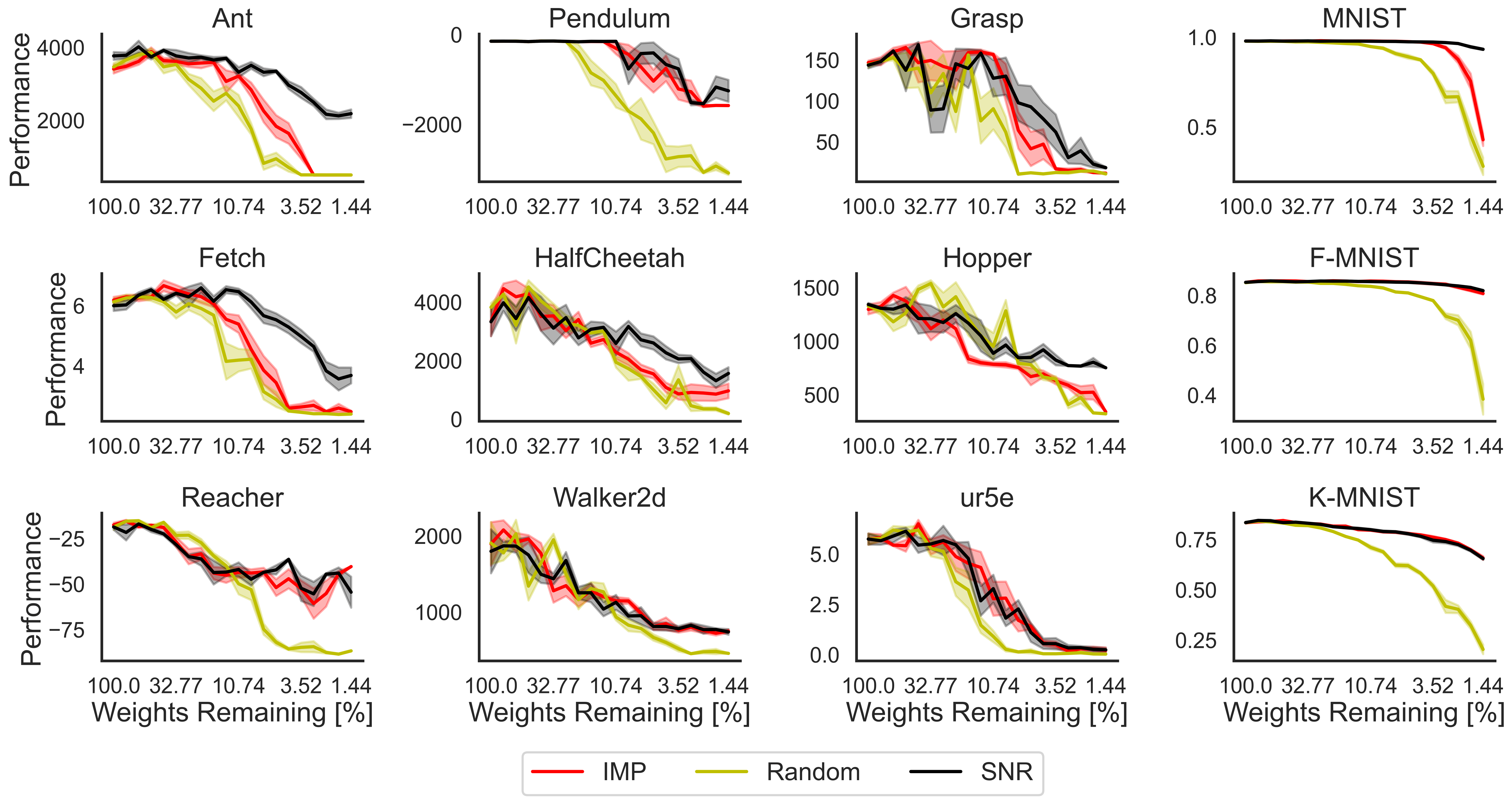}%
\caption{Existence of sparse evolvable intializations and benefits of SNR-based pruning. For the majority of considered settings we observe a lottery ticket effect (\textit{IMP} vs.\ \textit{random} pruning).
Furthermore, the proposed SNR-based iterative pruning consistently discovers initializations that outperform IMP tickets across all sparsity level. This highlights the positive effect of accounting for loss curvature at the time of pruning.
The size of the ticket effect depends on the task-network-ES setting, which indicates a difference in task-specific degree of overparametrization. \textbf{Top.} Sep-CMA-ES-based \citep{ros2008simple} tickets. \textbf{Middle.} SNES-based \citep{schaul2011high} tickets. \textbf{Bottom.} PGPE-based \citep{sehnke2010parameter} tickets. Results are averaged over 5 independent runs \& we plot standard error bars.}
\label{fig:existence}
\end{figure*}

\section{Winning Lottery Ticket Initializations Exist for Evolutionary Optimization}
\label{sec3:existence}

Previous work \citep{evci2020gradient, maene2021towards} suggests that GD ticket initializations overcome the sparsity-induced decrease in gradient flow by biasing initializations to retrain to the same loss basin. 
Given that ES are prone to higher stochasticity during the learning process, it is likely to converge to more diverse loss basins, raising the question whether the LTH phenomenon is unique to GD training using backpropagation. 

In order to investigate this question, we start by exhaustively evaluating the existence of winning lottery tickets in ES. 
We focus on 12 tasks, different network architectures and 3 ES: 
First, we evolve MLP agents (1 to 3 layers) on 9 Reinforcement Learning tasks focusing on the continuous control setting. We evolve the agents to maximize episodic returns using Sep-CMA-ES, SNES and PGPE. The environments are implemented by the Brax \citep{freeman2021brax} package and the agent evaluation is done using the average return on a set of test episodes. 
Next, we evolve CNN architectures on both the standard MNIST, Fashion-MNIST (F-MNIST) and Kuzushiji-MNIST (K-MNIST) digit classification tasks and to minimize the cross-entropy loss. 
In this case we evaluate the performance via the test accuracy of the final mean search statistic $\theta_{G, t}$.
For each task-network-ES combination we run the ES-adapted pruning procedure and prune at each iteration 20 percent of the remaining non-pruned weights ($p=0.2$). To ensure trainability at higher levels of sparsity, we generously calibrated the amount of ES generations and refer the reader to the SI \ref{sec:SI_hyperparameters} for a detailed overview of the considered hyperparameters.

\textbf{Winning lottery tickets consistently exist for various ES}. We find that the magnitude-based lottery ticket configuration outperforms the random-reinitialization baseline across the majority of tasks, network architectures and ES combinations (Figures \ref{fig:conceptual} and \ref{fig:existence}, red versus yellow curves). While the qualitative effect is robust across most settings, the quantitative magnitude differs significantly across task complexities and the degree of network overparametrization: For the simpler classification and pendulum task the observed ticket effect is large compared to the more complex control tasks such as HalfCheetah and the Hopper task. This indicates a relationship between task complexity and the overparametrization required to achieve high performance.

\begin{figure*}
\centering
\includegraphics[width=0.95\textwidth]{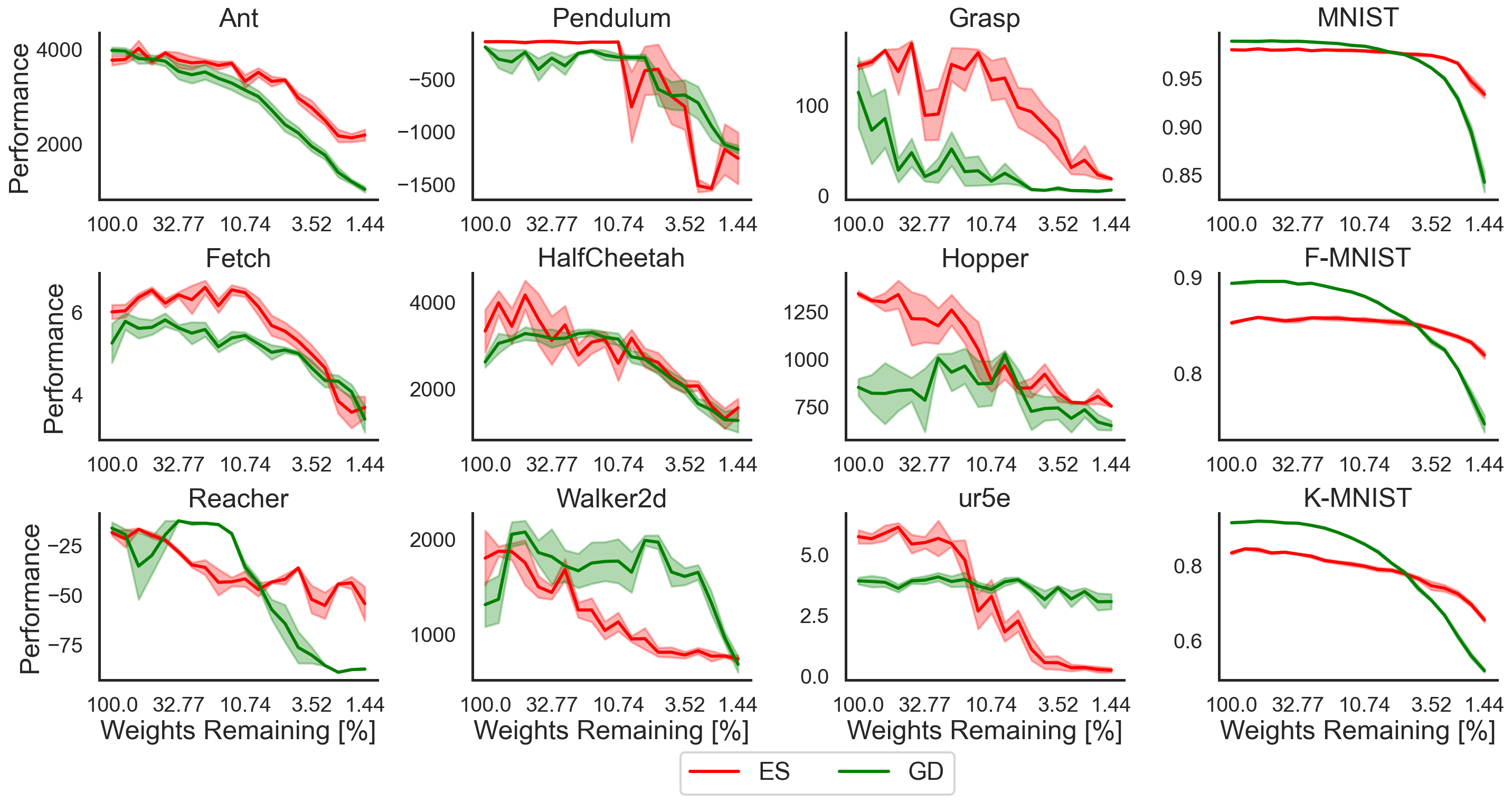}%
\caption{Performance comparison between GD and ES-based lottery ticket initializations. For the majority of considered task settings, ES-based training performs on par with GD for medium levels of sparsity. At high sparsity degrees, on the other hand, ES-SNR initializations tend to outperform the sparsity-matched GD initializations. \textbf{Top.} Sep-CMA-ES-based \citep{ros2008simple} tickets. \textbf{Middle.} SNES-based \citep{schaul2011high} tickets. \textbf{Bottom.} PGPE-based \citep{sehnke2010parameter} tickets. Results are averaged over 5 independent runs \& we plot standard error bars.}
\label{fig:es_sgd}
\end{figure*}

\textbf{SNR-based IMP results in sparser trainable tickets.} Next, we compare ES lottery tickets resulting from magnitude- and SNR-based pruning (Figures \ref{fig:conceptual} and \ref{fig:existence}, black versus red curves). We find that SNR-derived tickets are consistently trainable at higher degrees of sparsity. The search standard deviation of a specific weight indirectly incorporates information about the local loss landscape geometry.
Intuitively, weights with high associated standard deviation imply a strong robustness to perturbations and in turn a flat direction in the loss surface. Hence, we may expect weights with relatively small SNR to have a negligible performance contribution.

\textbf{Highly sparse trainability can be achieved by ES}. In Figures \ref{fig:conceptual} and \ref{fig:es_sgd} (green versus red curves) we compare the performance of GD-based IMP and ES-based SNR pruning. We find that sparse ES initializations are capable of outperforming GD-based training methods on several tasks. More importantly, we observe that ES-based methods are robustly better at training highly sparse networks. For the control tasks, ES can also outperform GD for moderate levels of sparsity (e.g. Hopper and Grasp tasks). For the vision-based tasks, on the other hand, GD-IMP starts to degrade in performance faster as the sparsity increases. In Section \ref{sec5:analysis} we investigate these dynamics and relationship between the sharpness of GD-based local optima and sparsity.



In summary, winning lottery tickets can be identified for different gradient-free ES, tasks and architectures. The size of the observed ticket effect and hence sparse trainability can be improved by considering not only the mean magnitude, but also the covariance as a reflection of loss curvature. The general phenomenon of sparse trainability of neural networks, therefore, is not unique to gradient descent and generalizes to algorithms with inherently higher stochasticity. The effect size and underlying contributions are task and network sensitive but largely robust to the optimization scheme. Finally, ES consistently train to higher performance at high sparsity levels.

\begin{figure*}
\centering
\includegraphics[width=\textwidth]{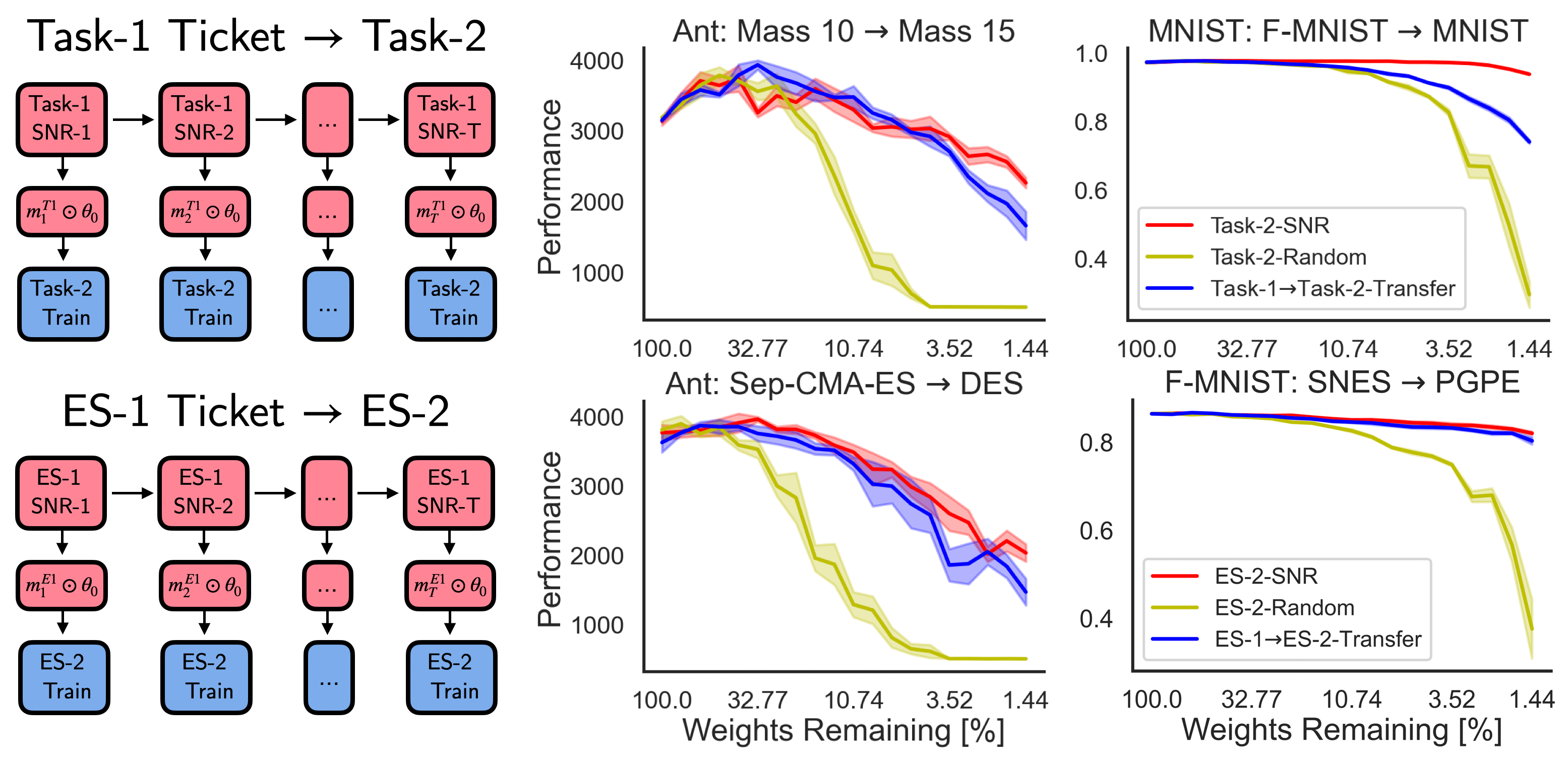}%
\caption{ES ticket initializations transfer between related tasks and between different ES. \textbf{Left.} Conceptual visualization of ticket initialization transfer between tasks/ES. We first run SNR-pruning on a task/ES setting to obtain a set of initializations at different sparsity levels. Afterwards, we evaluate their trainability in a different setting. \textbf{Top.} Task transfer between two torso masses (Ant task; Sep-CMA-ES) and image classification tasks (MNIST variants; SNES). \textbf{Bottom.} ES Transfer between Sep-CMA-ES and DES (Ant task) as well as SNES and PGPE (F-MNIST task). For both cases we can observe a positive transfer of the previously discovered pruning masks. Results are averaged over 5 independent runs \& we plot standard error bars.}
\label{fig:transfer_tasks}
\end{figure*}

\section{ES-Based Lottery Tickets Transfer across Tasks, ES and to GD-Based Training}
\label{sec4:transfer}

\begin{table}
\begin{tabular}{ |p{2cm}||p{2.15cm}|p{2.25cm}| }
 \hline \hline
 Task/ES & Pruning-Train & Transfer-Eval  \\
 \hline
\textit{Sep-CMA-ES} & Ant Mass 10 & Ant Mass 15 \\
\textit{SNES} & F-MNIST & MNIST \\
\hline
\textit{Ant} & Sep-CMA-ES    & DES \\
 \textit{F-MNIST} & SNES  &  PGPE \\
 \hline
\textit{Ant} & Sep-CMA-ES    & PPO \\
 \textit{F-MNIST} & SNES  &  SGD \\
 \hline \hline
\end{tabular}
\caption{Lottery Ticket ES/Task Transfer Settings}
\label{table1:transfer}
\end{table} 

The iterative pruning-based generation of winning tickets is a costly procedure requiring the sequential evolution of network initializations at different sparsity levels. Thereby, it remains impractical for most real world application. But is it possible to amortize the costly ticket generation for one task by successfully applying it to a different one?
Previous work showed that the GD-IMP-derived input layer mask captures task-specific information \citep{vischer2021lottery}, discarding task irrelevant dimensions.
We wondered whether ES tickets extract similar useful transferable inductive biases or whether they overfit to their generating setting (i.e., task, ES algorithm or training paradigm). If they remain transferable, we can hope for a task-general application of sparse ticket initializations in the context of ES.
To answer this question, we test the transferability of sparse initializations generated with ES-based SNR pruning to new unseen but related tasks.
We take inspiration by the work of \citet{morcos_2019} and re-train sparse initializations generated for one task-ES configuration on a different setting with a shared network architecture.\\
We start by examining the transfer of SNR-based initializations between different but related task variants and consider several settings (Figure \ref{fig:transfer_tasks}, top). Importantly, the source and transfer task share the same input/output dimensionality and are related (Table \ref{table1:transfer}, top; different torso mass for Ant control and digit/cloth image classification transfer).

\begin{figure*}
\centering
\includegraphics[width=\textwidth]{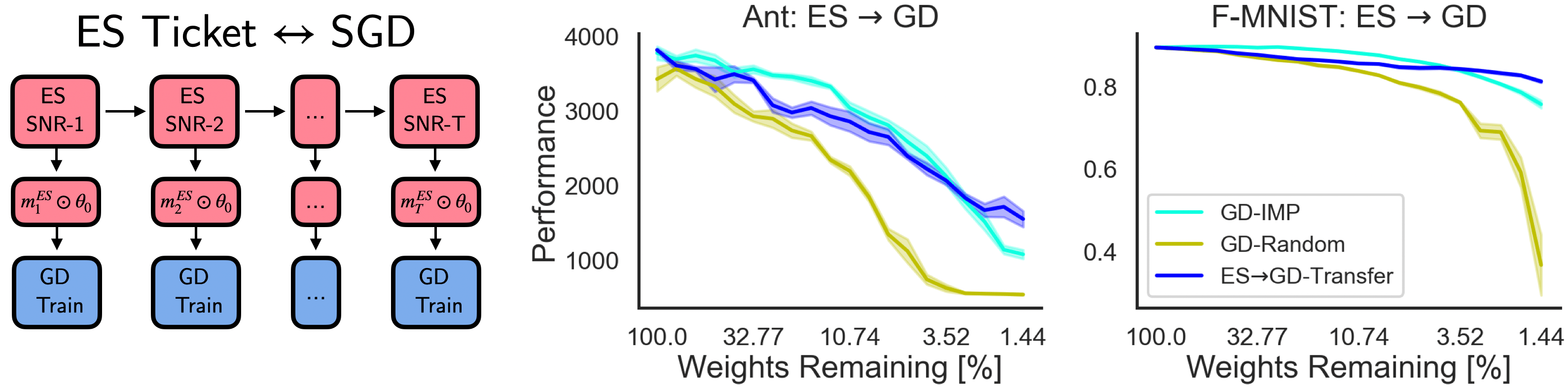}%
\caption{Transferability of ES tickets to GD-based training. \textbf{Left.} Conceptual visualization of ticket transfer between tasks. We first run SNR-based pruning using ES to obtain a set of initializations at different sparsity levels. Afterwards, we evaluate their trainability with GD. \textbf{Middle/Right.} For the two considered task-network-ES settings (Ant task and Sep-CMA-ES, F-MNIST and SNES), we observe a positive transfer effect between the two training paradigms when compared to a random pruning baseline. Furthermore, the ES initializations transfer their trainability at high levels of sparsity over to the GD training setting. The results are averaged over 5 independent runs \& we plot standard error bars.}
\label{fig:transfer_opt}
\end{figure*}

\textbf{Winning ES tickets can be transferred to related tasks}. The transfer of a ticket initialization derived on a related task improves the performance of a network trained on a new `sibling' task. The effect measured by the performance difference of the transferred initializations (blue) and the random pruning baseline (yellow) is significant across both of the considered tasks. The transferred ticket does not outperform the task-native SNR-lottery ticket (at high sparsity), indicating that tickets tend to capture both task-general and task-specific information. This positive transfer effect can be observed both for control (MLP) and vision (CNN) tasks. 

Next, we investigated whether it is possible to evolve sparse initializations generated by one evolution strategy with a different optimization strategy. We consider the transfer within the class of finite difference-based ES, to a covariance matrix adaptation-style ES and to GD-based training.


\textbf{Winning ES tickets can be transferred between ES}. Ticket initializations also transfer well between different ES optimization algorithms (Figure \ref{fig:transfer_tasks}, bottom). Oftentimes the transferred initializations train to the same performance level of the ES-specific ticket initialization, indicating that a within-task transfer is easier as compared to the previous across-task setting. This observation again holds for both task settings (control and vision) as well as different ES combinations.

\textbf{Winning ES tickets can be transferred to GD Training.} Finally, we repeat the procedure from the previous subparagraphs, but this time transfer sparse ES-derived initializations to downstream training with GD. Again, we find a positive effect for transferring an initialization that was obtained by ES (Figure \ref{fig:transfer_opt}).
As discussed in Section \ref{sec3:existence}, ES tickets can perform worse than GD-based training for moderate levels of sparsity. In line with this observation, we find that the size of the transfer effect correlates with the relative performance differences between the two paradigms. We do not find a strong positive effect for sparsity levels where the GD ticket baseline outperforms the ES ticket (e.g.\ for the ant task). More interestingly, for very sparse networks the ES-transferred initialization can even outperform the GD-ticket indicating that highly sparse ES pruning masks are well transferable to GD training. 

\section{Linear Mode Connectivity \& SNR Pruning}
\label{sec5:analysis}

\begin{figure*}
\centering
\includegraphics[width=0.95\textwidth]{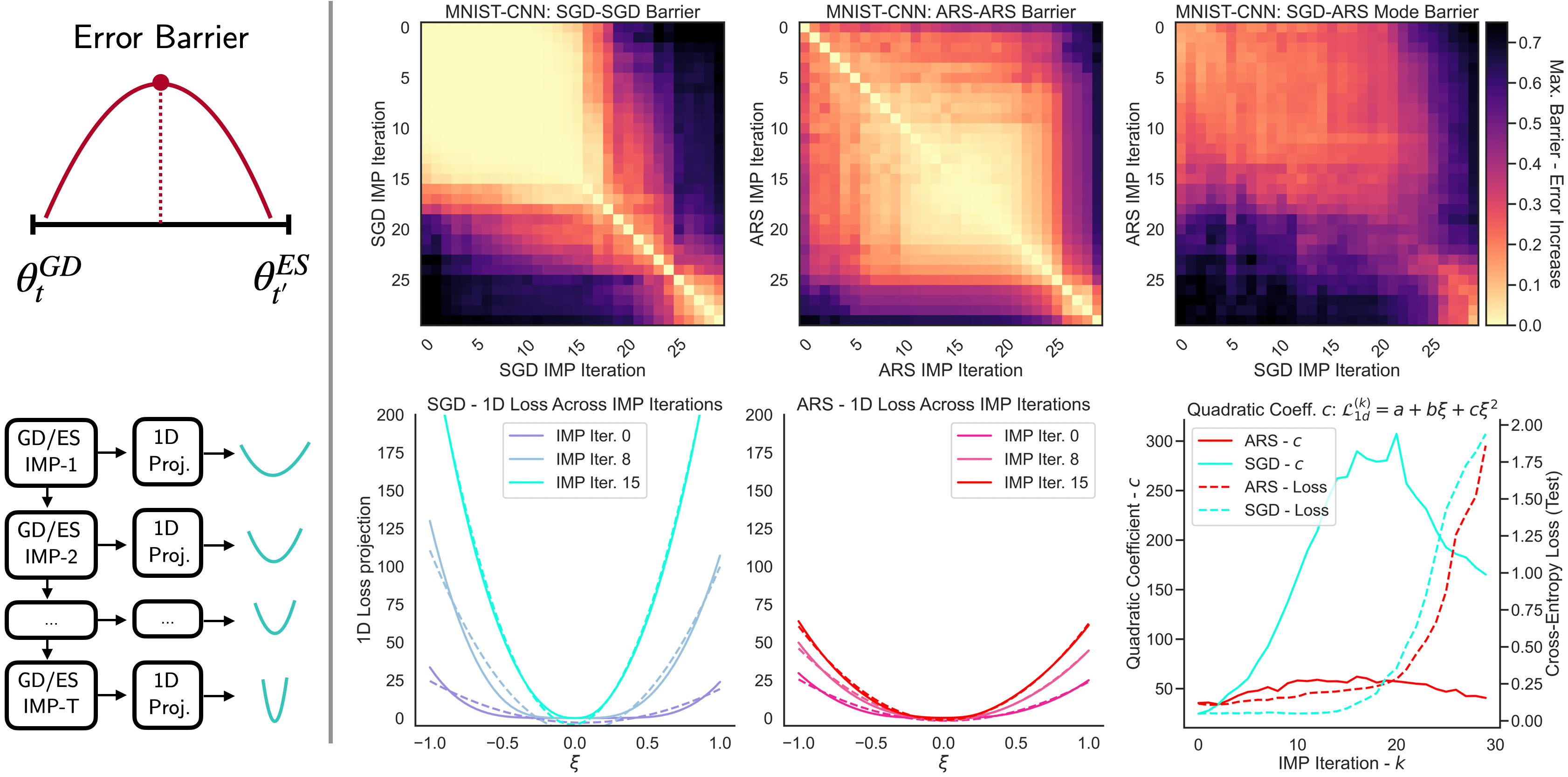}%
\caption{Connectivity barriers \& local minima sharpness in GD and ES \citep[ARS,][]{mania2018simple} on MNIST. \textbf{Top.} 
While for low sparsity GD-based optima can be linearly connected with nearly no drop in test accuracy, ARS-based optima suffer from small barriers. GD and ARS optima cannot be connected without strong performance drops, indicating that they find different optima. \textbf{Bottom.} Low-dimensional loss projections \citep{li2017simple} and curvature estimates at different iterations. The sharpness of GD-based optima increases rapidly with the sparsity level. ES-based optima remain more flat throughout the IMP procedure. Results are averaged over 5 independent runs.}
\label{fig:mode_connect}
\end{figure*}

\textbf{ES and GD optima are not linearly connected.} Based on the previous results, we wondered how the trained models obtained by ES and GD differed. Therefore, we compared the linear connectivity \citep{frankle_2020b} of the local optima across sparsity levels. We compute the test accuracy ($\mathcal{A}$) error barrier between two trained networks $\max_{\alpha \in [0, 1]} \mathcal{A}(\alpha \theta + (1 - \alpha) \theta')$ for a range of $\alpha$, comparing network combinations at different IMP iterations.
In line with previous work \citep{frankle_2019b}, GD-based local optima remain strongly connected for moderate levels of sparsity (Figure \ref{fig:mode_connect}, top). ES-based \citep[ARS,][]{mania2018simple} solutions, on the other hand, already experience a performance barrier between early IMP iterations, but remain better connected at higher sparsity levels. The optima found by GD and ES are generally not linearly connectable, indicating that GD and ES find fundamentally different solutions.
Furthermore, it questions a generalization of the regurgitating ticket interpretation to ES \citep{evci2020gradient, maene2021towards}: ES ticket initializations exist despite the fact that they do not repeatedly train to the same loss basin.  
In SI Figure \ref{fig:supp_weights_mnist}, we find that the lack of ES-GD connectivity can partially be explained by different weight magnitudes for the two training paradigms. In general, GD-based solutions have higher magnitude weights and tend to prune the input layer less.

\textbf{ES tends to converge to flatter optima.} A natural follow-up question is: How do the curvatures of local optima obtained by the different training paradigms differ? We use one-dimensional random projections \citep{li2018visualizing} of the test loss $\mathcal{L}(\theta; \xi) = \mathcal{L}(\theta + \xi \eta)$ with $\eta \sim \mathcal{N}(\mathbf{0}, \mathbf{I})$ to examine the sensitivity of the discovered local optimum for different strengths $\xi \in [-1, 1]$. We quantify the approximate curvature by fitting a quadratic polynomial to the projected loss as a function of the perturbation strength. In Figure \ref{fig:mode_connect} (bottom) we observe that the approximate sharpness of the GD optima increases rapidly with the sparsity level. For ES-based optima, on the other hand, the curvature increases at a smaller rate across IMP iterations. We provide visualizations of the 2D projected loss surface in SI Figure \ref{fig:supp_2d_mnist}.

\begin{figure}
\centering
\includegraphics[width=0.45\textwidth]{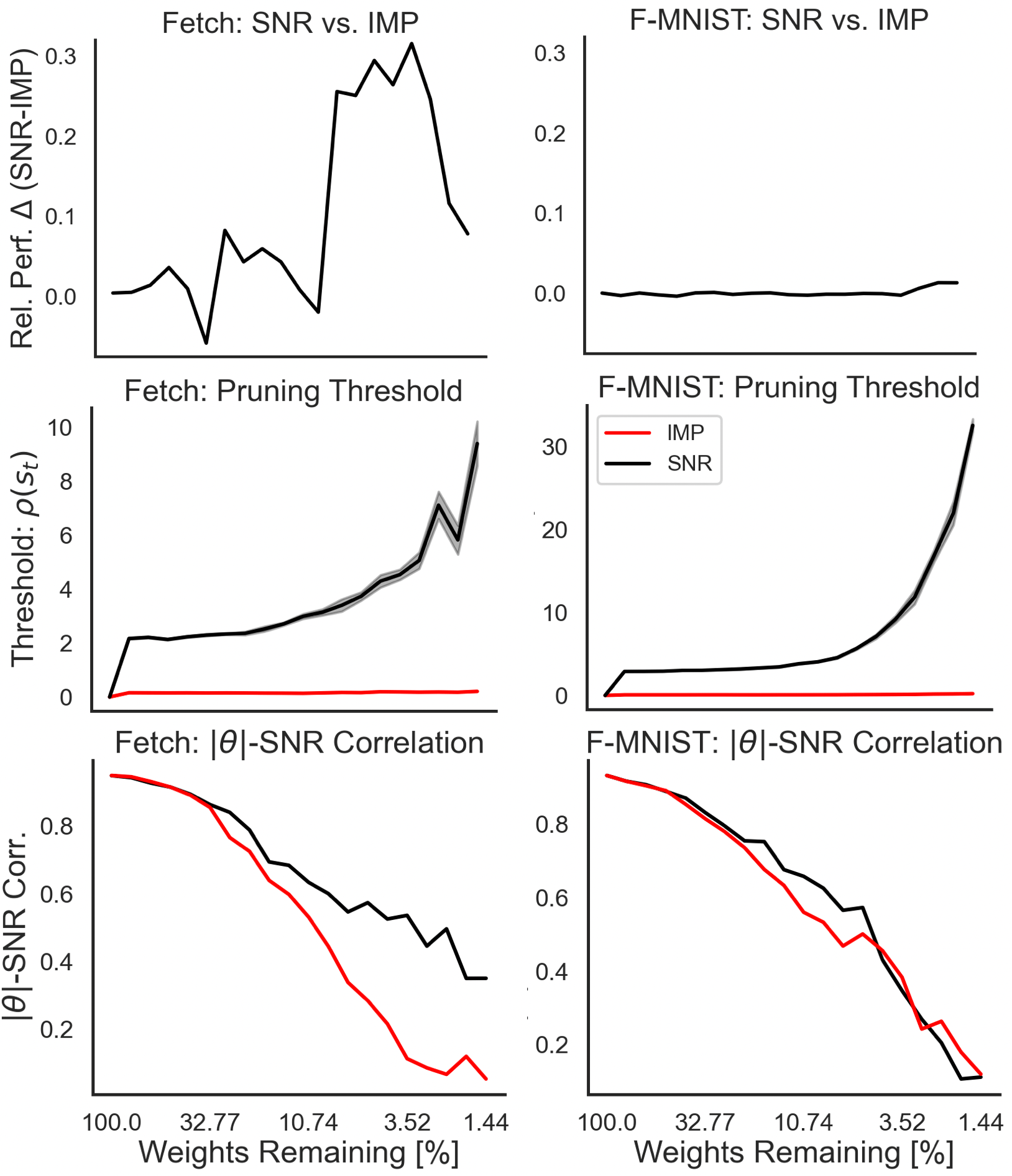}%
\caption{\textbf{Top.} Relative performance difference between IMP and SNR pruning for Fetch \& F-MNIST tasks. \textbf{Middle.} Pruning thresholds for IMP \& SNR. \textbf{Bottom.} Correlation between SNRs and weight magnitudes across pruning iterations. The correlation remains high in the case of a positive performance difference. The results are averaged over 5 independent runs.}
\label{fig:snr_pruning}
\end{figure}

\textbf{SNR pruning dynamically accounts for fitness curvature.} Finally, we investigate conditions under which SNR-based pruning improves over IMP. In Figure \ref{fig:snr_pruning} we plot the correlation of weight magnitudes and SNRs across pruning stages. The correlation decreases for both SNR and IMP-based pruning runs as sparsity increases. The SNR-IMP performance gap is closely related to the relative correlation dynamics: If the correlation decreases faster for IMP than SNR (left; Fetch task), one also observes a positive impact of SNR pruning on the performance. Otherwise, we do not (right; F-MNIST task). This indicates that SNR-based pruning can account for non-isotropic changes in the fitness landscape curvature caused by sparsification. Dimensions with high sharpness (small $\sigma$) will have a larger SNR, which makes them less prone to pruning. Future work will have to uncover the mechanistic underpinnings of this phenomenon.

\newpage
\section{Discussion}

\textbf{Summary.} We establish the existence of sparse trainable initializations in evolutionary optimization. Sparse trainability, therefore, does not require a specific masked gradient flow. The exact size of the ticket effect depend on ES, architecture and task. The resulting sparse initializations are transferable across training paradigms and to related tasks. Tickets in ES do not necessarily retrain to the same loss basin but still remain trainable across sparsity levels. \\
\textbf{Ethical Considerations.} 
\citet{hooker2020characterising} show that compression can amplify bias and decrease fairness in the GD-based training setting. 
As we scale ES, future work will have to assess whether these problems transfer to ES model compression and how to mitigate them. \\
\textbf{Limitations.} This work is limited by its empirical nature and scalability of ES. Furthermore, our study focus on medium network sizes. This is partially due to ES suffering from a lack of hyperparameter understanding and the adoption of tools tailored towards GD-based optimization (optimizers,  etc.). Finally, our analysis is based on the computationally costly iterative pruning procedure, which requires multiple sequential training runs.\\
\textbf{Future Work.} 
Dynamic sparse training with ES provides a direction for future work and may enable protocols which simultaneously grow and prune networks. Furthermore, a full understanding of sparse trainability requires a theoretical treatment of the effect of sparsity on the fitness landscape.


\newpage
\bibliography{main}
\bibliographystyle{icml2022}

\newpage
\appendix
\onecolumn
\parttoc 

\section{Additional Results}

\subsection{Lottery Ticket Baseline Comparison for GD-Based IMP}

\begin{figure}[H]
\centering
\includegraphics[width=0.85\textwidth]{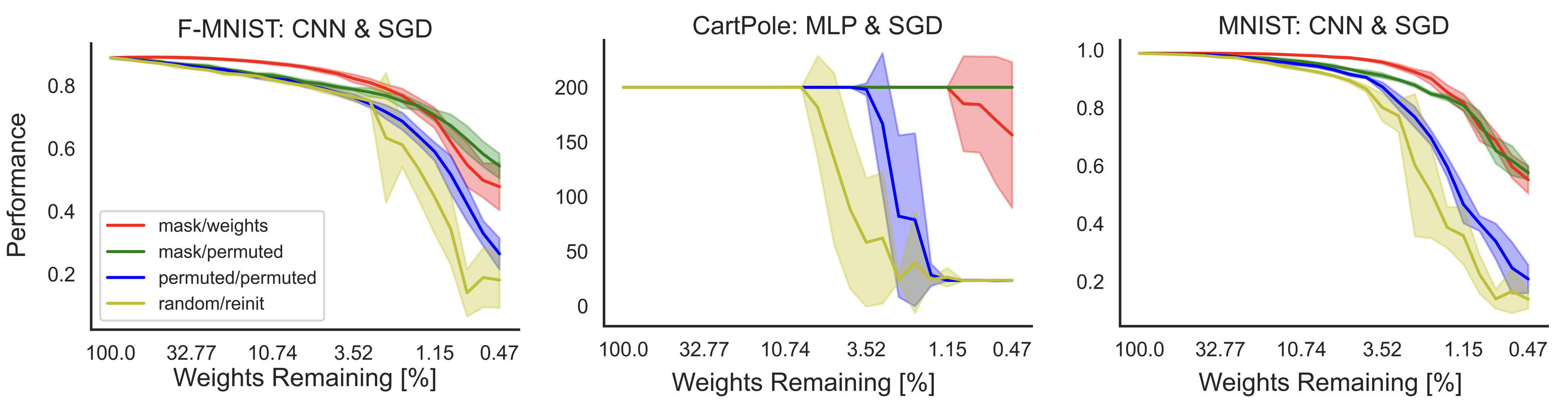} \\%
\includegraphics[width=0.85\textwidth]{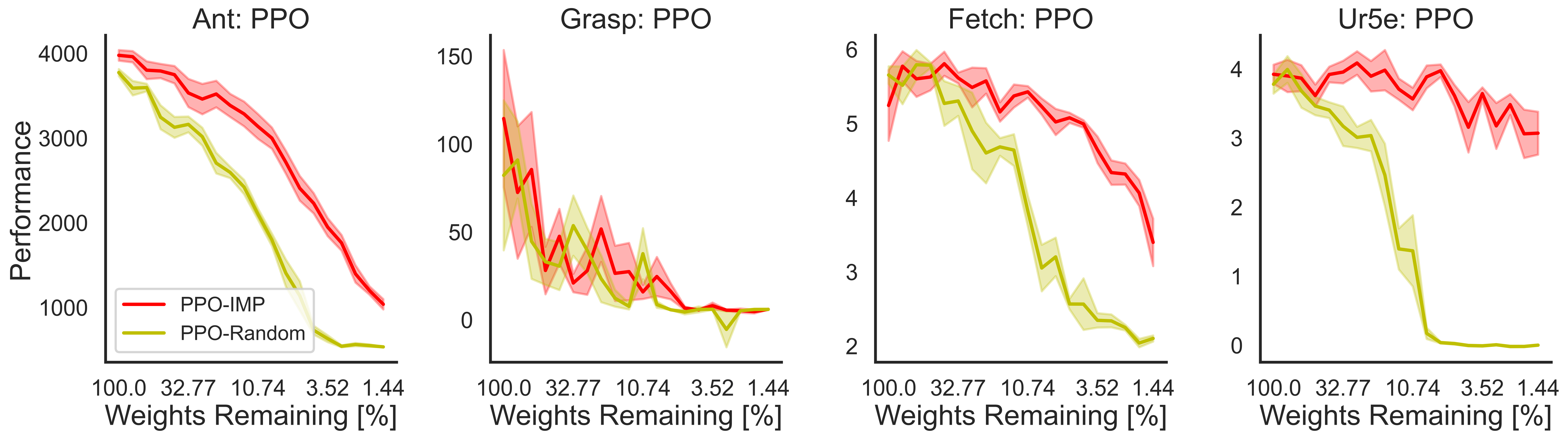}%
\caption{GD (Adam) baseline comparison \citep{vischer2021lottery}. The task and the architecture dictate the importance of mask, preserved weights and layerwise pruning ratio. The results are averaged over 5 independent runs and we plot one standard deviation intervals.}
\label{fig:supp_sgd_blines}
\end{figure}

\subsection{Impact of Network Initialization}
\vspace{-0.25cm}
\begin{figure}[H]
\centering
\includegraphics[width=0.75\textwidth]{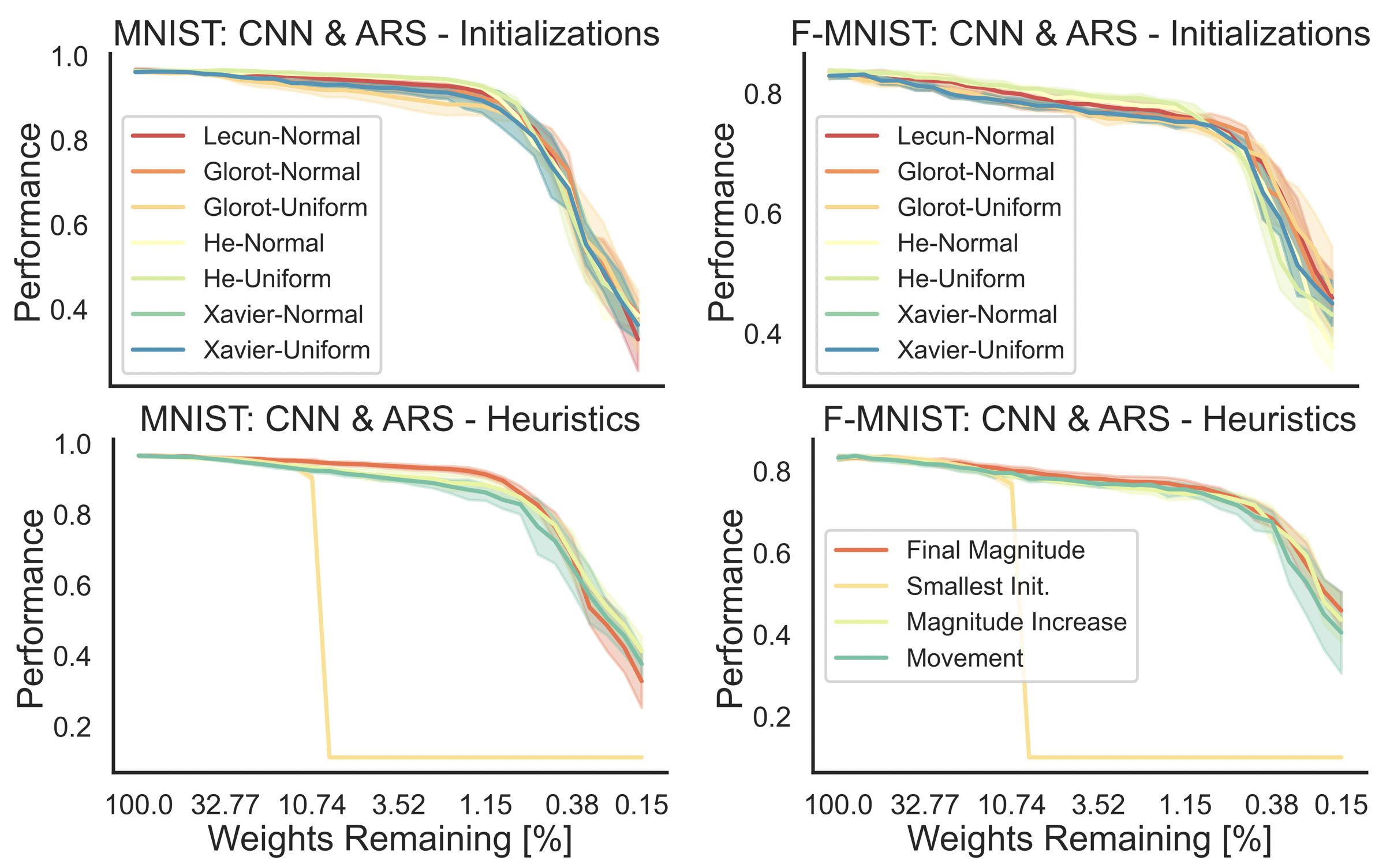}%
\caption{\textbf{Top}. Network initialization comparison. We validate whether the chosen network initialization has a significant influence on sparse evolvability. Across the four tasks and 3 ES we do not observe any significant differences between the considered initialization schemes. \textbf{Bottom}. Pruning heuristic comparison. We follow \citet{zhou_2019} and compare 4 different pruning heuristics: Final magnitude ($|\theta_f|$), movement ($|\theta_f - \theta_0|$), magnitude increase ($|\theta_f| - |\theta_0|$) and smallest initial value ($\theta_0$). Across the four tasks and 3 ES we can observe that all of the heuristics which depend on $\theta_f$ provide sensible choices.  The results are averaged over 5 independent runs and we plot one standard deviation intervals.}
\label{fig:supp_initialization}
\end{figure}




\subsection{2D Loss Projections}

Following \citet{li2018visualizing} we provide 2D visualizations of the loss surface using random projections and for different IMP iterations. We sample two random vectors $\eta_1, \eta_2 \sim \mathcal{N}(0, I)$. Afterwards, we perform a filter-/neuron-wise normalization of the random vectors based on the corresponding norms of the trained dense network at IMP iteration 0. The same normalized random vectors are then applied to the optima found after training at different IMP iterations $\theta_f$: $\mathcal{L}(\theta_f; \alpha, \beta) = \mathcal{L}(\theta_f + \alpha \eta_1 + \beta \eta_2)$. We evaluate the loss on the test set using $\alpha, \beta \in [-1, 1]$ for a discretized range (51 steps). Afterwards, we apply spline smoothing to obtain the heat visualizations.

\begin{figure}[H]
\centering
\includegraphics[width=0.75\textwidth]{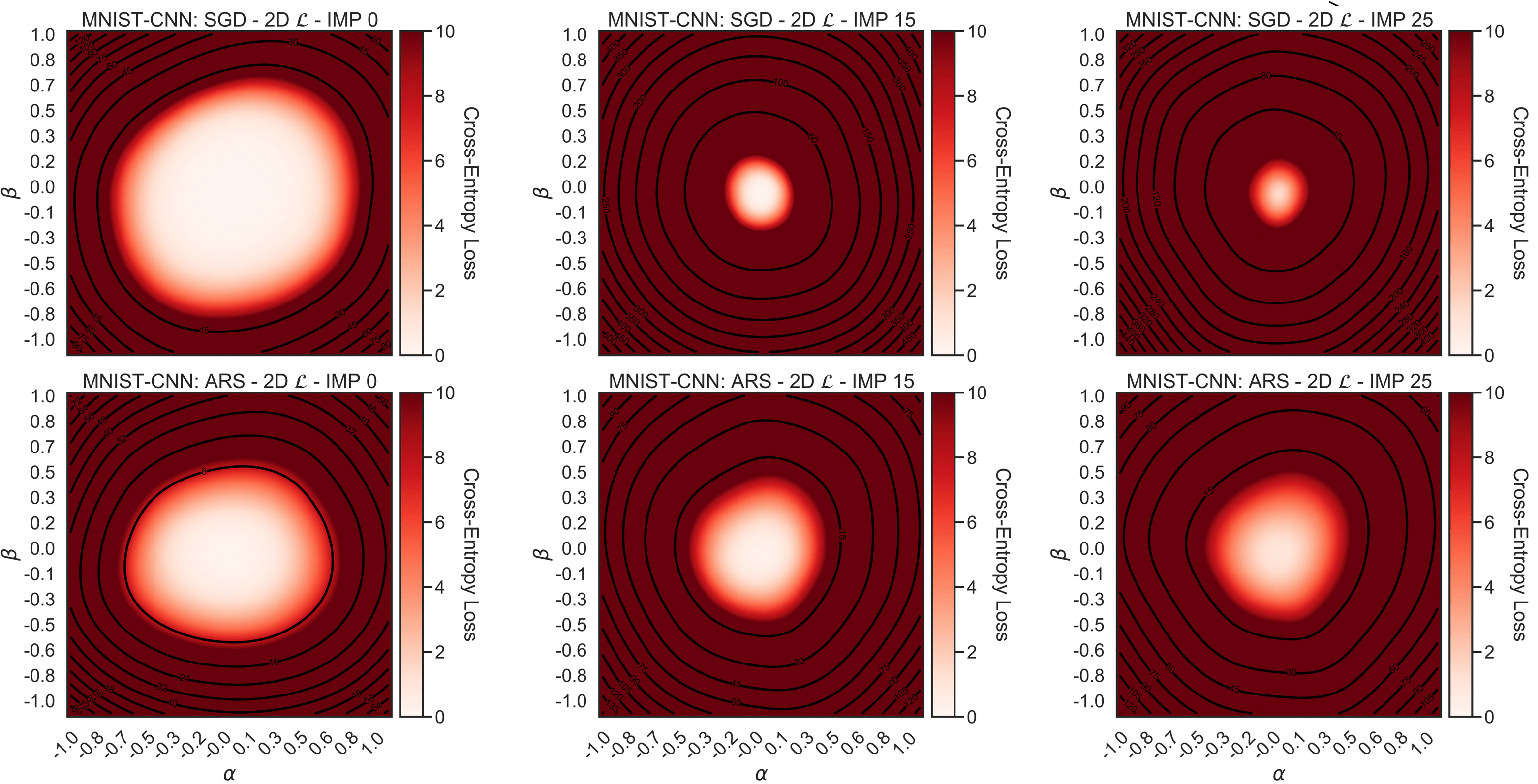}%
\caption{2D loss projection for MNIST task \citep{li2018visualizing}. GD optima become sharper as sparsity increases, while ES optima remain flat. The results are averaged over 5 independent runs.}
\label{fig:supp_2d_mnist}
\end{figure}


\subsection{Layerwise Pruning Ratios and Weight Summary Statistics}

\begin{figure}[H]
\centering
\includegraphics[width=0.8\textwidth]{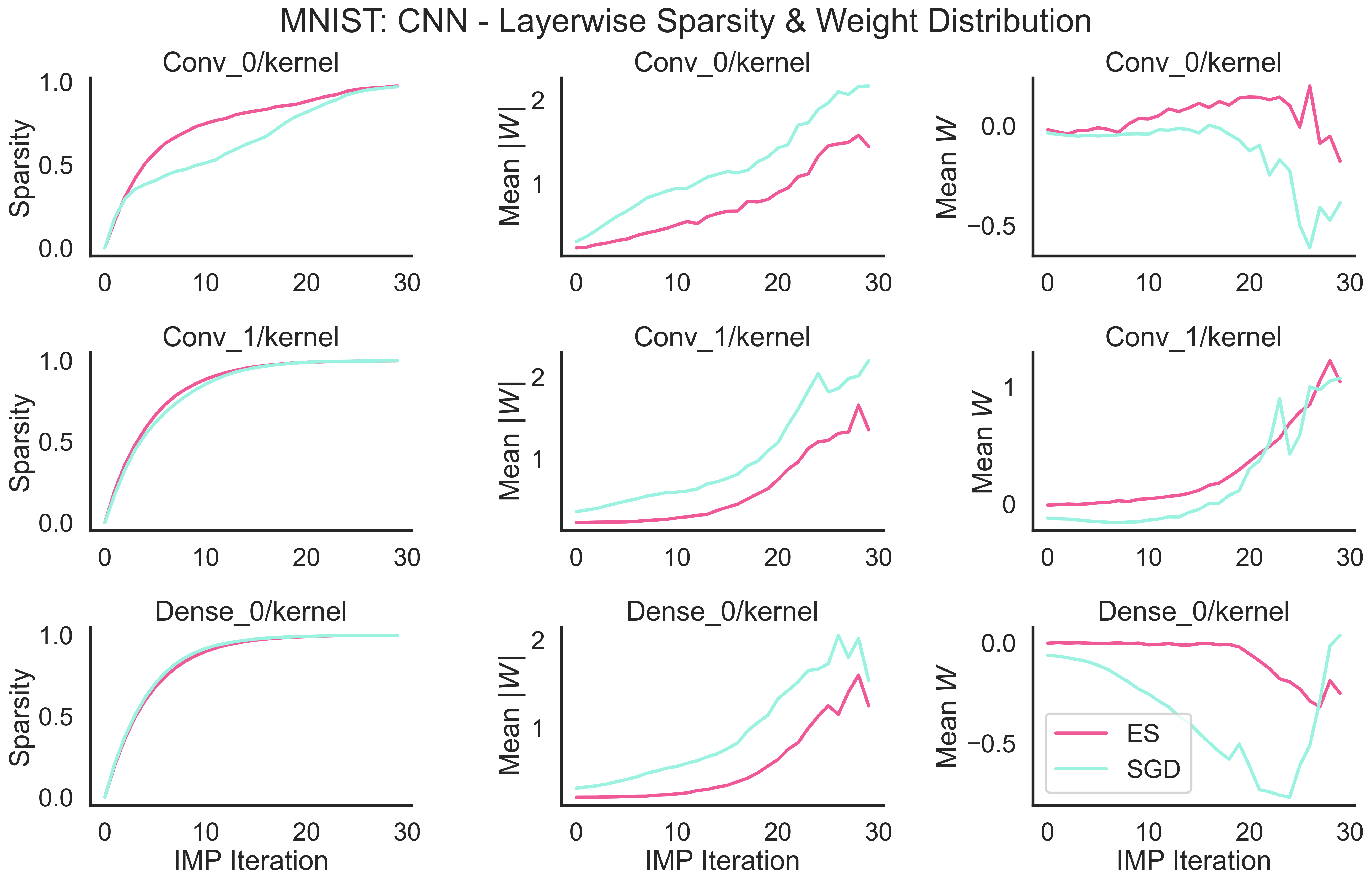}%
\caption{Layerwise pruning ratios and weight summary statistics for MNIST task. The results are averaged over 5 independent runs.}
\label{fig:supp_weights_mnist}
\end{figure}



\section{Software \& Compute Requirements}
\label{sec:compute}

Simulations were conducted on a high-performance cluster using 5 or more independent runs (random seeds). Depending on the setting, a single iterative pruning run (20 individual training runs at different sparsity levels) lasts between 60 minutes and 10 hours of training time:

\begin{itemize}
    \item Brax tasks \& MLP policy: 1 NVIDIA V100S GPU, ca. 10 hours. 
    \item Pendulum task \& MLP network: 1 NVIDIA RTX 2080Ti GPU, ca. 1 hour.
    \item MNIST/F-MNIST/K-MNIST task \& CNN network: 1 NVIDIA RTX 2080Ti GPU, ca. 2 hours.
\end{itemize}

The code is publicly available under \url{https://github.com/RobertTLange/es-lottery}.
The experiments were organized using the \texttt{MLE-Infrastructure} \citep[MIT license]{mle_infrastructure2021github} training management system. 
All training loops and ES are implemented in JAX \citep{jax2018github}. 
All visualizations were done using Matplotlib \citep[]{hunter_2007} and Seaborn \citep[BSD-3-Clause License]{waskom_2021}. Finally, the numerical analysis was supported by NumPy \citep[BSD-3-Clause License]{harris_2020}. Furthermore, we used the following libraries:

\begin{itemize}
    \item Evosax: \citet{evosax2022github}, Gymnax: \citet{gymnax2022github}, Evojax: \citet{evojax2022}, Brax: \citet{freeman2021brax}
\end{itemize}

Note that we focus our experiments and analysis on Evolution Strategies and do not study Genetic Algorithms \citep[e.g.][]{rechenberg1978evolutionsstrategien, clune2008natural, kumar2022effective, lange2023discovering}. They do not maintain an explicit search distribution and instead use an archive of parent solutions to apply selection and mutation rate adaptation. Future work may investigate pruning the network architecture based on the best parent member after running the Genetic Algorithm.

\section{Hyperparameter Settings}
\label{sec:SI_hyperparameters}

\subsection{Control - Shared MLP Task Hyperparameters}

\vspace{-0.25cm}
\begin{table}[H]
\parbox{.35\linewidth}{
\centering
\small
\begin{tabular}{|r|l||r|l|}
\hline \hline
Parameter & Value & Parameter & Value \\ 
\hline
Population                 & 256   & \# Generations        & 3k - 4k      \\
\# Train Eval                & 8   & \# Test Eval        & 164      \\
Network Type                & MLP   & \# Hidden Units        & 32-80      \\
Output Layer           & Tanh   & \# Seeds        & 5      \\
Objective           & Return   &   Evaluation     &  Return     \\
ES Optimizer           & Adam   &   \# Hidden Layers     &  1-3     \\
\hline \hline
\end{tabular}
\caption{ES Settings for Control MLP tasks.}}
\hfill
\parbox{.55\linewidth}{
\centering
\small
\begin{tabular}{|r|l||r|l|}
\hline \hline
Parameter & Value & Parameter & Value \\ 
\hline
Env Steps                 & 40-1200 Mio   & Obs Norm        & True      \\
\# Train Eval                & 8   & \# Test Eval        & 164      \\
Network Type                & MLP   & \# Hidden Units        & 32-80      \\
Output Layer           & Tanh   & \# Seeds        & 5      \\
Objective           & PPO   &   Evaluation     &  Return     \\
Optimizer           & Adam   &   \# Hidden Layers     &  1-3     \\
\hline \hline
\end{tabular}
\caption{PPO Settings for Control MLP tasks.}
}
\end{table}

For PPO Brax tasks we used the default settings provided in the repository \citep{freeman2021brax}.

For ES Brax tasks we used the settings provided by \citep{lange2022discovering}.

\subsection{Vision - Shared CNN Task Hyperparameters}

\vspace{-0.25cm}
\begin{table}[H]
\parbox{.45\linewidth}{
\centering
\small
\begin{tabular}{|r|l||r|l|}
\hline \hline
Parameter & Value & Parameter & Value \\ 
\hline
Population                 & 128   & \# Generations        & 4000      \\
Batch Size                & 1024   & Filter sizes     &  [5, 5]     \\
Network Type                & CNN   & \# Conv Layers        & 2      \\
Output Layer           & Dense   & \# Seeds        & 5     \\
Objective           & CE   &   Evaluation     &  Acc     \\
ES Optimizer           & Adam   &   Conv Filters     &  [8, 16]     \\
\hline \hline
\end{tabular}
\caption{ES Settings for MNIST CNN.}}
\hfill
\parbox{.45\linewidth}{
\centering
\small
\begin{tabular}{|r|l||r|l|}
\hline \hline
Parameter & Value & Parameter & Value \\ 
\hline
\# Epochs                 & 20   & Optimizer        & Adam      \\
Batch Size                & 128   & Learning rate        & 3e-04      \\
Network Type                & CNN   & \# Conv Layers        & 2      \\
Output Layer           & Dense   & \# Seeds        & 5      \\
Objective           & CE   &   Evaluation     &  Acc     \\
Filter sizes     &  [5, 5]   &   Conv Filters     &  [8, 16]     \\
\hline \hline
\end{tabular}
\caption{SGD Settings for MNIST CNN.}
}
\end{table}
\vspace{-0.25cm}

\end{document}